\documentclass[sigconf]{acmart}

\usepackage{multirow}
\usepackage{makecell}
\usepackage{algorithm}
\usepackage{algpseudocode}
\usepackage{amsthm} 
\newtheoremstyle{defstyle} 
{15pt} 
{15pt} 
{\itshape} 
{\parindent} 
{\bfseries} 
{.} 
{.5em} 
{} 

\theoremstyle{defstyle}

\usepackage{graphicx}
\usepackage{subcaption}
\usepackage{tabularx}
\usepackage{array}
\usepackage{float}

\AtBeginDocument{%
  \providecommand\BibTeX{{%
    \normalfont B\kern-0.5em{\scshape i\kern-0.25em b}\kern-0.8em\TeX}}}

\renewcommand\footnotetextcopyrightpermission[1]{}

\begin{document}

\title{FedCAda: Adaptive Client-Side Optimization for Accelerated and Stable Federated Learning}

\author{Liuzhi Zhou}
\authornote{Both authors contributed equally to this research.}
\affiliation{%
  \institution{School of Computer Science, Fudan University}
  \city{Shanghai}
  \country{China}
}
\email{liuzhizhou22@m.fudan.edu.cn}

\author{Yu He}
\authornotemark[1]
\affiliation{%
  \institution{School of Computer Science, Fudan University}
  \city{Shanghai}
  \country{China}
}
\email{yuhe22@m.fudan.edu.cn}

\author{Kun Zhai}
\affiliation{%
  \institution{School of Computer Science, Fudan University}
  \city{Shanghai}
  \country{China}
}
\email{kdi22@m.fudan.edu.cn }

\author{Xiang Liu}
\affiliation{%
  \institution{Tandon School of Engineering, New York University}
  \city{New York}
  \country{USA}
}
\email{xl493@nyu.edu}

\author{Sen Liu}
\affiliation{%
  \institution{School of Computer Science, Fudan University}
  \city{Shanghai}
  \country{China}
}
\email{senliu@fudan.edu.cn}

\author{Xingjun Ma}
\affiliation{%
  \institution{School of Computer Science, Fudan University}
  \city{Shanghai}
  \country{China}
}
\email{xingjunma@fudan.edu.cn}

\author{Guangnan Ye}
\authornote{Corresponding author.}
\affiliation{%
  \institution{School of Computer Science, Fudan University}
  \city{Shanghai}
  \country{China}
}
\email{yegn@fudan.edu.cn}

\author{Yu-Gang Jiang}
\affiliation{%
  \institution{School of Computer Science, Fudan University}
  \city{Shanghai}
  \country{China}
}
\email{ygj@fudan.edu.cn}

\author{Hongfeng Chai}
\affiliation{%
  \institution{School of Computer Science, Fudan University}
  \city{Shanghai}
  \country{China}
}
\email{hfchai@fudan.edu.cn}



\begin{abstract}
Federated learning (FL) has emerged as a prominent approach for collaborative training of machine learning models across distributed clients while preserving data privacy. However, the quest to balance acceleration and stability becomes a significant challenge in FL, especially on the client-side. In this paper, we introduce FedCAda, an innovative federated client adaptive algorithm designed to tackle this challenge. FedCAda leverages the Adam algorithm to adjust the correction process of the first moment estimate $m$ and the second moment estimate $v$ on the client-side and aggregate adaptive algorithm parameters on the server-side, aiming to accelerate convergence speed and communication efficiency while ensuring stability and performance. Additionally, we investigate several algorithms incorporating different adjustment functions. This comparative analysis revealed that due to the limited information contained within client models from other clients during the initial stages of federated learning, more substantial constraints need to be imposed on the parameters of the adaptive algorithm. As federated learning progresses and clients gather more global information, FedCAda gradually diminishes the impact on adaptive parameters. These findings provide insights for enhancing the robustness and efficiency of algorithmic improvements. Through extensive experiments on computer vision (CV) and natural language processing (NLP) datasets, we demonstrate that FedCAda outperforms the state-of-the-art methods in terms of adaptability, convergence, stability, and overall performance. This work contributes to adaptive algorithms for federated learning, encouraging further exploration.
\end{abstract}





\maketitle

\section{Introduction}

Federated learning (FL) \cite{mcmahan2017communication}, as a popular distributed machine learning architecture, uploads aggregated updates to the server instead of raw data. Federated learning trains a global model by iteratively training models locally on each client using local data and then aggregating them on the server, allowing clients to collaboratively train while preserving privacy, resulting in a global model encompassing information from all clients, typically outperforming models trained on individual client data. Federated learning algorithms are currently classified into two scenarios \cite{kairouz2021advances}: cross-silo and cross-device. Cross-silo involves a few large institutions engaging in federated learning, each possessing substantial datasets, such as consumer transaction records from different e-commerce platforms. Cross-device entails federated learning across multiple devices, where the number of devices is large but the data on each device is small; however, participation of all devices in every round cannot be guaranteed, as in the case of edge devices and the Internet of Things (IoT).  The most commonly used federated learning algorithm is FedAvg \cite{mcmahan2017communication}, also known as local SGD and Parallel SGD, which ensures stable utilization of vanilla stochastic gradient descent (SGD) by each client locally, even with minimal communication rounds, while maintaining proximity to the global model. However, in practical applications, several challenges are often encountered, such as 1) significant communication overhead due to slow convergence leading to repetitive communication between clients and the server, and 2) lack of adaptability, as fixed-step-size SGD updates are not conducive to handling heavy-tailed statistical gradient noise distributions, which typically occur in large-scale model training tasks such as ViT \cite{dosovitskiy2020image}, GPT-3 \cite{brown2020language}, GANs \cite{goodfellow2020generative}, and especially given the outstanding performance of large models across various tasks nowadays.

In order to address the aforementioned challenges, adaptive gradient methods such as AdaGrad \cite{duchi2011adaptive}, RMSProp \cite{tieleman2017divide}, Adam \cite{kingma2014adam}, and AMSGrad \cite{reddi2019convergence} have been widely embraced in centralized learning, surpassing SGD in hyperparameter tuning and convergence speed in deep learning. However, the direct application of these methods to federated learning poses significant challenges. Existing integrations of adaptive gradients and federated learning primarily occur on the server-side, exemplified by FedAdaGrad \cite{reddi2020adaptive} and its variants, FedAMS\cite{wang2022communication}, among others. When employing adaptive methods on the client-side, challenges arise due to issues such as client drift, stemming from factors like local data biases, leading to convergence difficulties. Limited research has explored client-side local adaptation, including the use of AMSGrad in a local context, termed local-AMSGrad, the incorporation of adaptive gradient methods into federated learning such as FAFED \cite{wu2023faster}, and the application of Stochastic Polyak Step-size (SPS) method in the federated learning setting, known as FedSps \cite{mukherjee2023locally}. Consequently, achieving efficient adaptive federated learning while ensuring convergence remains a compelling research problem.

\begin{figure*}[htbp]
\centering
\includegraphics[width=\linewidth]{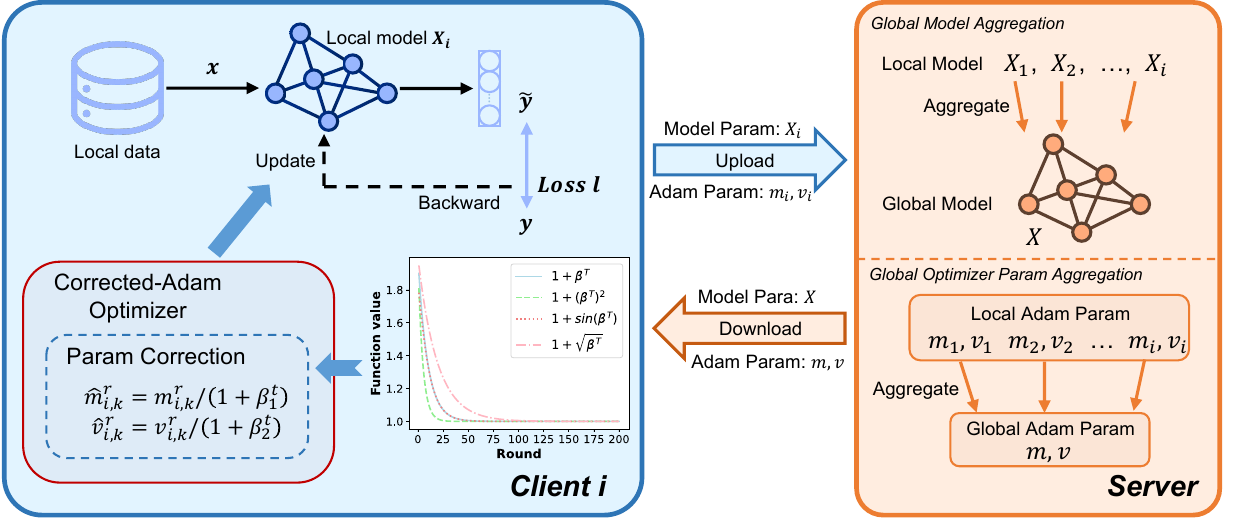}
\caption{Overview of the proposed client-side adaptive federated learning (FedCAda). Right: The server takes on two roles: \textcircled{1} Aggregate the model weights from the clients and distribute them to each model for the next round; \textcircled{2} Aggregate the Adam optimizer parameters $m$ and $v$ from the clients, and similarly distribute the aggregated average parameters to each client for the next round. Left: The clients utilize local data for training, wherein, during the parameter update stage of backward, the adjusted-Adam optimizer is used in place of the traditional SGD optimizer in FedAvg to achieve client-side adaptive optimization.}
\label{overview}
\end{figure*}

Utilizing adaptive gradient methods on the client-side to accelerate convergence speed and communication efficiency presents a challenge in balancing acceleration with stability. Excessive acceleration may jeopardize the stability of federated learning algorithms, causing local models to accelerate toward their respective directions without constraint mechanisms. Therefore, it's crucial to consider the relationship between client-side model adaptation and the global model. We aim to ensure that client-side models accelerate convergence through adaptive gradients while preventing them from deviating excessively from the global model, thus avoiding convergence issues. To address this, we propose a novel approach for FL. In this paper, we specifically design a federated client adaptive algorithm, FedCAda, which applies the Adam algorithm to the clients in federated learning. Our model architecture is illustrated in Figure \ref{overview}. This algorithm focuses on the adjustment process of $m$ and $v$ while aggregating adaptive algorithm parameters on the client-side. By achieving client-side adaptation in federated learning, FedCAda accelerates convergence speed while maintaining better stability and improving overall performance. Additionally, we explore the use of various functions to optimize the first moment estimate $m$ and the second moment estimate $v$ adjustment algorithms. In summary, our contributions primarily include:
\begin{itemize}
\item We emphasize the importance of balancing acceleration and stability when applying adaptive algorithms on federated learning clients. We propose FedCAda, a federated client adaptive algorithm that achieves better stability and performance while accelerating federated learning clients.
\item We investigate the impact of modifying the denominator during the adjustment process. Specifically, we compare the original subtraction method with three alternative functions: exponential function, power function, and trigonometric function. This comparative analysis reveals the effectiveness of different approaches towards zero.
\item We conduct a series of experiments on CV and NLP datasets and compare with a large number of state-of-the-art (SOTA) adaptive algorithms. The experiments demonstrate that FedCAda outperforms SOTA methods in terms of adaptability, convergence, stability, etc.
\end{itemize}

\section{Related Work}
\subsection{FedAvg-Based Federated Learning} 
Federated learning, as an emerging paradigm in machine learning, aims to facilitate joint model training while ensuring data privacy. It has garnered significant attention in the machine learning community. FedAvg \cite{mcmahan2017communication}, as the pioneering classic federated learning algorithm, establishes a standard optimization approach. By periodically averaging local SGD updates, it effectively reduces communication overhead and mitigates key issues in federated learning, such as client data heterogeneity and partial client participation to some extent. However, in scenarios with severely heterogeneous data sets (non-IID), local model deviations can occur, causing the central model to stray from the global optimum, resulting in performance discrepancies compared to independently identically distributed (IID) data \cite{zhao2018federated,long2023fedcd,qi2023cross}. Therefore, several approaches, based on FedAvg, have been proposed to address client heterogeneity issues by introducing regularization terms such as FedProx \cite{li2020federated}, FedDyn \cite{jin2023feddyn}, SCAFFOLD \cite{karimireddy2020scaffold}, FFReID \cite{zhang2023improving} among others. FedProx penalizes the distance between server and client parameters by incorporating a proximal term. FedDyn dynamically adjusts local objectives to ensure the stability point convergence of local optima and global objectives. SCAFFOLD employs control variables to correct client drift in their local updates. FFReID introduces a proximal term and a feature regularization term for local model training to improve local training accuracy and global aggregation convergence. Additionally, some approaches focus on incorporating momentum from the server side into clients \cite{qu2022generalized}, utilizing contrastive learning loss functions \cite{li2021model}, pre-training \cite{nguyen2022begin} or knowledge distillation \cite{lin2020ensemble,zhang2023cuing, lu2023federated} to improve client drift. FedSAM \cite{caldarola2022improving}, for instance, employs the SAM optimizer on clients to converge towards flat minima.

\subsection{Adaptive Federated Learning}
Adaptive methods are one of the most significant variants in machine learning distinct from stochastic gradient descent. They encompass optimization algorithms such as Adam, AdaGrad, and AdaDelta. These methods are commonly employed in training deep neural networks, and in many cases, they exhibit superior convergence speed compared to SGD or other methods. Subsequently, several variants of these methods have been extensively studied. AMSGrad, for instance, addresses convergence issues in Adam and introduces improvements.

In the context of federated learning, the application of adaptive methods necessitates careful consideration of the trade-off between convergence stability and acceleration. Regarding server-side optimization, Reddi et al. \cite{reddi2020adaptive} pioneered the use of adaptive optimizers in the federated learning setting, introducing a series of adaptive FL methods such as FedAdagrad, FedYogi, and FedAdam. Wang et al. \cite{wang2022communication} further proposed FedAMS based on FedAdam and introduced FedCAMS with communication compression, along with improved convergence analysis. On the client-side optimization front, Chen et al. \cite{chen2023convergence} introduced the Local AMSGrad algorithm, noting that using adaptive methods directly on clients can lead to convergence issues, unlike on the server-side. Recently, Wu et al. \cite{wu2023faster} built upon this work and proposed an adaptive method based on momentum-based variance reduction techniques, effectively reducing communication complexity. Additionally, some approaches achieve adaptivity by directly adjusting the step size of client updates.

\section{Method}
\subsection{Problem Statement}
In this paper, we possess a total of $N$ clients, each client holding its own local data $x_i$ distributed according to $\mathcal{D}_i$, labeled by $y_i$. Our objective is to investigate the following non-convex optimization problem in federated learning:
\begin{equation}
    \min_{x \in \mathbb{R}^d} f(\mathbf{x}) = \frac{1}{N} \sum_{i=1}^{N} \mathcal{F}_i(\mathbf{x}), 
\end{equation}
where $d$ denotes the dimension of the model parameters, $\mathcal{F}_i(\mathbf{x}) := \mathbb{E}_{\xi \sim \mathcal{D}_i} \mathcal{L}_i(\mathbf{x}; \xi_i)$ is the local non-convex loss function on client $i \in [N]$, where $\theta_i$ denotes the parameters of local model, whose structure is assumed to be identical across the clients and the server. Since we mainly focus on the non-i.i.d data setting, where local datasets have heterogeneous distributions, $\mathcal{D}_i$, $\mathcal{D}_j$ can vary from each other, i.e., $\mathcal{D}_i \neq \mathcal{D}_j, \forall i \neq j$.

To address the optimization issue mentioned above, FedAvg adopts a periodic averaging of locally trained SGD. At round $r$, the global model $x_t$ is dispatched to participating clients. Clients train locally using their own data, executing $k$ steps of SGD updates to optimize the model, then transmitting the model differences, termed as $\delta_i^r$, to the server. The server updates the global model by averaging the pseudo-gradients received from the clients.

\begin{figure}[htbp]
\centerline{\includegraphics[width=\linewidth]{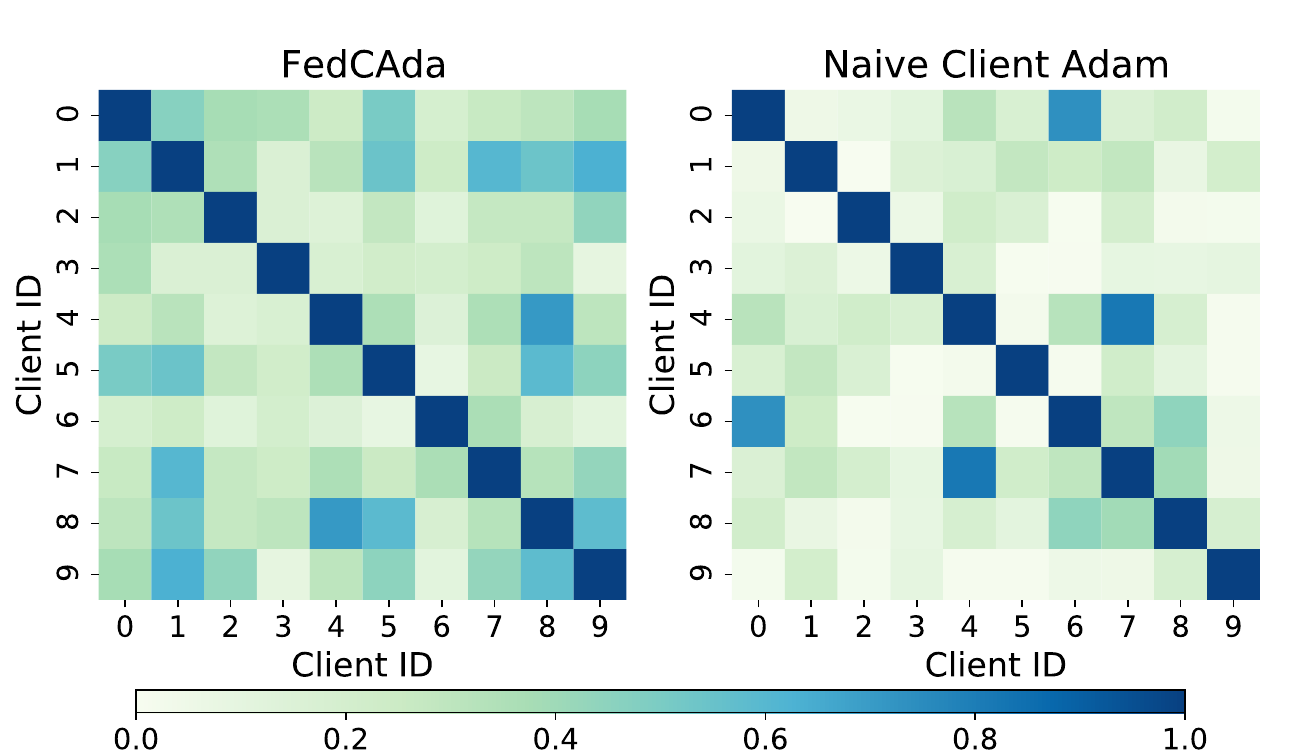}}
\caption{The CKA similarity of the first moment estimate $m$ of the Adam optimizer after 200 training rounds among 10 clients. (CKA outputs a similarity score between 0 and 1, indicating not similar at all to identical)}
\label{fig1}
\end{figure}

While applying adaptive methods, such as Adam, to the clients in federated learning, the most straightforward approach is to replace the local SGD optimizer in the FedAvg framework with the Adam optimizer, as depicted in Algorithm \ref{alg1}. However, although the local SGD optimizer ensures the convergence of the global model in the FedAvg framework, replacing it with the Adam optimizer introduces convergence issues. This issue arises from the application of adaptive methods in the federated learning framework, which causes the adaptive learning rates of different clients to diverge from each other. We computed the Centered Kernel Alignment (CKA) to measure the similarity of the first moment estimate $m$ and the second moment estimate $v$ of the Adam optimizers after 200 training rounds among 10 clients. In Figure \ref{fig1}, we illustrate the CKA similarity of the first moment estimate $m$ of the Adam optimizer (the second moment estimate $v$ is similarly depicted). It can be observed that using the vanilla method, the parameters in the Adam optimizers across different clients demonstrate lower similarity. This implies that under the vanilla method, the adaptive learning rates across clients exhibit higher heterogeneity, leading to convergence issues. \\

\textit{\textbf{CKA Similarity}: Centered Kernel Alignment (CKA) \cite{kornblith2019similarity} is a technique used to assess the similarity between the output features of two distinct neural networks, given the same input dataset. Initially, a dataset $D_{CKA}$ of size $N$ is selected, from which feature matrices are extracted. One matrix, denoted as $Z_1$, is derived from the output of the first neural network, possessing dimensions $N \times d_1$. Simultaneously, another matrix, $Z_2$, is obtained from the output of the second neural network, with dimensions $N \times d_2$. To prepare these feature matrices for comparison, a preprocessing step known as centering is conducted, involving the centering of columns within each matrix. Following this preprocessing, the linear CKA similarity between the two representations, $X$ and $Y$, can be computed. This computation is facilitated by the formula: }
\begin{equation}
CKA(X, Y) = \frac{\|Z_1^TZ_2\|_F^2}{\|Z_1^TZ_1\|_F^2 \|Z_2^TZ_2\|_F^2}.
\end{equation}

\begin{algorithm}
\caption{Vanilla Client Adaptive Federated Learning Algorithm}\label{alg1}
\begin{algorithmic}[1]
\Require initial parameters $\mathbf{x}_i^0 = \mathbf{x}^0$; learning rates $\eta_l$ and $\eta_g$; $\beta_1; \ \beta2; \ \epsilon$.
\State $m_{i,0}^1 \leftarrow 0$; $v_{i,0}^1 \leftarrow 0$
\For{round $r = 1, \dots, T$}
\State Server samples a subset of clients $\mathcal{S}_t^r$.
\State $\mathbf{x}_{i,0}^r=\mathbf{x}^r$
\For{each client $i \in \mathcal{S}_t^r$ in parallel}
\For{local step $k = 0, \dots, K - 1$}
\State Compute a local training estimate $\tilde{g}_{i,k}^r =$
\Statex \qquad \qquad \quad$\nabla F(\mathbf{x}_{i,k}^r;\xi_{i,k}^r)$
\State $//$ FedAvg employs local SGD:
\State $\mathbf{x}_{i,k}^r = \mathbf{x}_{i,k-1}^r - \eta_l \tilde{g}_{i,k}^r$
\State $//$ Utilize Adam as a substitute for local SGD:
\State $m_{i,k}^r = \beta_1 m_{i,k-1}^r + (1-\beta_1) \tilde{g}_{i,k}^r$
\State $v_{i,k}^r = \beta_2 v_{i,k-1}^r + (1-\beta_2) [\tilde{g}_{i,k}^r]^2$
\State $\hat{m}_{i,k}^r = m_{i,k}^r / (1-\beta_1^t)$
\State $\hat{v}_{i,k}^r = v_{i,k}^r / (1-\beta_2^t)$
\State $\mathbf{x}_{i,k}^r = \mathbf{x}_{i,k-1}^r - \eta_l \frac{\hat{m}_{i,k}^r}{\sqrt{\hat{v}_{i,k}^r} + \epsilon}
$
\EndFor
\State $\Delta_i^r = \mathbf{x}_{i}^r - \mathbf{x}^r$
\EndFor
\State $\Delta^r = \frac{1}{|\mathcal{S}_t^r|} \sum_{i \in \mathcal{S}_t^r} \Delta_i^r$
\State $\mathbf{x}_{r+1} = \mathbf{x}_r + \eta_g \Delta^r$
\EndFor
\end{algorithmic}
\end{algorithm}

Hence, while implementing client acceleration, it is imperative to delve into the direction and magnitude of adaptive learning rates for each client, ensuring a more robust convergence. Subsequently, we integrate this notion into the design of our FedCAda algorithm, with the aim of achieving a stable adaptive optimizer that operates effectively on the client-side.

\begin{algorithm}
\caption{Federated Learning Client Adaptive (FedCAda) Algorithm}\label{alg2}
\begin{algorithmic}[1]
\Require initial parameters $\mathbf{x}_i^0 = \mathbf{x}^0$; learning rates $\eta_l$ and $\eta_g$; $\beta_1; \ \beta2; \ \epsilon$.
\State $m^1 \leftarrow 0$; $v^1 \leftarrow 0$
\For{round $r = 1, \dots, T$}
\State Server samples a subset of clients $\mathcal{S}_t^r$.
\State $\mathbf{x}_{i,0}^r=\mathbf{x}^r$, $m_{i,0}^r=m^r$, $v_{i,0}^r=v^r$
\For{each client $i \in \mathcal{S}_t^r$ in parallel}
\For{local step $k = 0, \dots, K - 1$}
\State Compute a local training estimate $\tilde{g}_{i,k}^r =$
\Statex \qquad \qquad \quad$\nabla F(\mathbf{x}_{i,k}^r;\xi_{i,k}^r)$

\State $m_{i,k}^r = \beta_1 m_{i,k-1}^r + (1-\beta_1) \tilde{g}_{i,k}^r$
\State $v_{i,k}^r = \beta_2 v_{i,k-1}^r + (1-\beta_2) [\tilde{g}_{i,k}^r]^2$
\State $\hat{m}_{i,k}^r = m_{i,k}^r / (1+\beta_1^t)$
\State $\hat{v}_{i,k}^r = v_{i,k}^r / (1+\beta_2^t)$
\State $\mathbf{x}_{i,k}^r = \mathbf{x}_{i,k-1}^r - \eta_l \frac{\hat{m}_{i,k}^r}{\sqrt{\hat{v}_{i,k}^r} + \epsilon}
$
\EndFor
\State $\Delta_i^r = \mathbf{x}_{i}^r - \mathbf{x}^r$
\EndFor
\State $m^{r+1} = \frac{1}{|\mathcal{S}_t^r|} \sum_{i \in \mathcal{S}_t^r} m_{i,k}^r$
\State $v^{r+1} = \frac{1}{|\mathcal{S}_t^r|} \sum_{i \in \mathcal{S}_t^r} v_{i,k}^r$
\State $\Delta^r = \frac{1}{|\mathcal{S}_t^r|} \sum_{i \in \mathcal{S}_t^r} \Delta_i^r$
\State $\mathbf{x}_{r+1} = \mathbf{x}_r + \eta_g \Delta^r$
\EndFor
\end{algorithmic}
\end{algorithm}

\subsection{FedCAda}
To address the convergence issues that may arise in vanilla client adaptive federated learning, we propose FedCAda. This method is based on vanilla client adaptive federated learning but modifies the structure of both the client and server sides.
\subsubsection{Server-side}
To address the heterogeneity issue of adaptive parameters across different clients, we introduced an additional aggregation of adaptive parameters for all clients selected in each round on the server side, alongside the original aggregation of model parameters. This unified the adaptive parameters, mitigating convergence difficulties caused by excessive heterogeneity of adaptive parameters.
\subsubsection{Client-side}
In the Adam algorithm, the process of correcting the adaptive parameters $m$ and $v$ is as follows:
\begin{equation}
  \hat{m}_{i,k}^r = m_{i,k}^r / (1-\beta_1^t)
\end{equation}
\begin{equation}
    \hat{v}_{i,k}^r = v_{i,k}^r / (1-\beta_2^t)
\end{equation}
We unified the adaptive parameters on the server-side during the aggregation rounds, but it is inevitable to still face issues caused by heterogeneity during the training process on the client-side. Therefore, we aim to further mitigate the impact of the heterogeneity of adaptive parameters on the client-side in our FedCAda algorithm. We adjust the correction process of $m$ and $v$ to limit their magnitudes, with the following formulas:
\begin{equation}
\label{equ5}
      \hat{m}_{i,k}^r = m_{i,k}^r / (1+\beta_1 ^{t})
\end{equation}
\begin{equation}
\label{equ6}
    \hat{v}_{i,k}^r = v_{i,k}^r / (1+\beta_2 ^{t})
\end{equation}

\begin{figure}[htbp]
\centerline{\includegraphics[width=\linewidth]{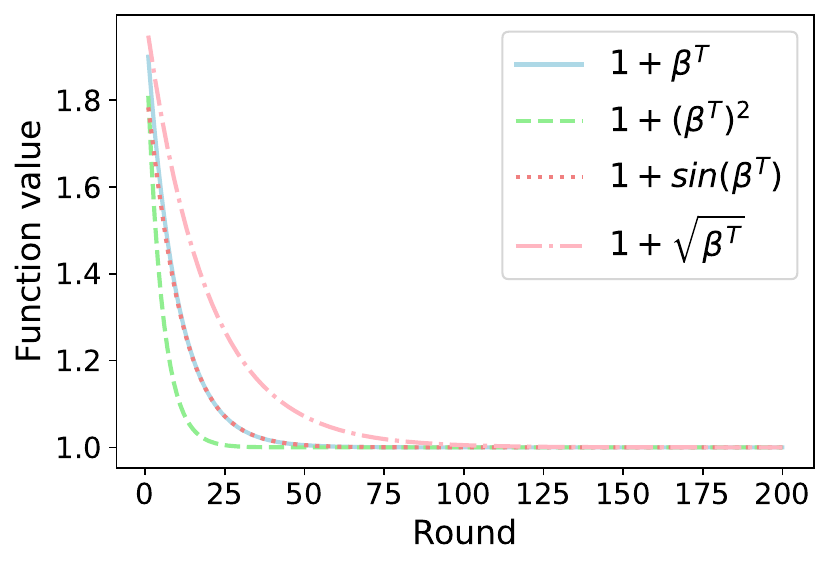}}
\caption{The curves of different functions used in adjustment functions under $\beta=0.9$ and $T=200$.}
\label{fig2}
\end{figure}

In the setting of Adam, both $\beta_1$ and $\beta_2$ are numbers greater than 0 and less than 1. As $T$ increases, the denominator in the above formulas tends to 1. Therefore, in our algorithm, FedCAda initially imposes a stronger limitation on the parameters of the adaptive algorithm due to the client-side models containing less information from other client models. Consequently, the adaptive learning rate decreases, compressing the update step size on the client side. As federated learning progresses and the client-side gathers more global information, FedCAda gradually reduces its impact on the adaptive parameters.

The pseudo-code of our comprehensive algorithm FedCAda is provided in Algorithm \ref{alg2}. Initially, $m^1$ and $v^1$ are initialized to $0$. At round $r$, the server selects a subset of clients $\mathcal{S}_t^r$. Then, each client $i \in \mathcal{S}_t^r$ initializes its own parameters using the model parameters $\mathbf{x}^r$ aggregated from the previous round on the server, along with the Adam optimizer parameters $m^r$ and $v^r$. The client obtains model $\mathbf{x}_{i,k}^r$ after performing $K$ steps of local iteration using the Adam optimizer, where the first moment estimate $m_{i,k}^r$ and the second moment estimate $v_{i,k}^r$ of the Adam optimizers are adjusted using equations (\ref{equ5}) and (\ref{equ6}) to $\hat{m}_{i,k}^r$ and $\hat{v}_{i,k}^r$. Finally, the server aggregates the $\Delta_i^r$ from each client as the pseudo gradient to update the global model and additionally aggregates the Adam optimizer parameters $m_{i,k}^r$ and $v_{i,k}^r$ from each client.

Besides, for the adjustment of $m_{i,k}^r$ and $v_{i,k}^r$ in Algorithm \ref{alg2}, lines 10 and 11, we provide four options. These options involve squaring, taking the sine, and square rooting of $\beta^t$, each offering a different adjustment rate. We simulated the trends of these adjustment functions over 200 rounds, assuming a value of $\beta = 0.9$. Their trend curves are illustrated in Figure \ref{fig2}.

\begin{align*}
      &\text{Option 1:} \quad \hat{m}_{i,k}^r = m_{i,k}^r / (1+\beta_1 ^{t}), \ \hat{v}_{i,k}^r = v_{i,k}^r / (1+\beta_2 ^{t}). \\
      &\text{Option 2:} \quad \hat{m}_{i,k}^r = m_{i,k}^r / (1+(\beta_1 ^{t})^2), \ \hat{v}_{i,k}^r = v_{i,k}^r / (1+(\beta_2 ^{t})^2). \\
      &\text{Option 3:} \quad \hat{m}_{i,k}^r = m_{i,k}^r / (1+\sin(\beta_1 ^{t})), \ \hat{v}_{i,k}^r = v_{i,k}^r / (1+\sin(\beta_2 ^{t})). \\
      &\text{Option 4:} \quad \hat{m}_{i,k}^r = m_{i,k}^r / (1+\sqrt{\beta_1 ^{t}}), \ \hat{v}_{i,k}^r = v_{i,k}^r / (1+\sqrt{\beta_2 ^{t}}). 
\end{align*}

\section{Experiments}
In this section, we evaluate the performance of FedCAda to highlight the efficacy of the proposed method in terms of performance, convergence speed, and robustness to data heterogeneity. To achieve this objective, we conducted simulations on three representative federated learning datasets and compared our approach against existing state-of-the-art (SOTA) federated learning methods, particularly adaptive federated learning algorithms, including FedAvg \cite{mcmahan2017communication}, FedAdam \cite{reddi2020adaptive}, FedAMS \cite{wang2022communication}, FedSps \cite{mukherjee2023locally}, and FaFed \cite{wu2023faster}. All experiments were primarily conducted utilizing the NVIDIA GeForce RTX 3090 Ti.

\subsection{Setup}

\textbf{Datasets and models.} We adhered to prior research and employed public benchmark datasets, encompassing two computer vision datasets for image classification, namely CIFAR-10 \cite{krizhevsky2009learning} and FashionMNIST \cite{xiao2017fashion}, as well as an NLP dataset for predicting the next character based on Shakespeare dataset \cite{mcmahan2017communication}. For IID data, we partitioned the dataset by randomly assigning training data to each client. For non-IID data, inspired by \cite{yurochkin2019bayesian}, we utilized a method involving Dirichlet distribution sampling within each training set of the datasets to simulate data heterogeneity, employing a Dirichlet distribution with $\alpha = 0.1$. Particularly, to better reflect real-world scenarios, clients not only possess categories with different distributions but also differ in dataset sizes \cite{li2023revisiting}. For CIFAR-10, we trained a SimpleCNN comprising three convolutional layers and two fully connected layers. For FashionMNIST, we trained an MLP, a three-layer MLP model with ReLU activation. As for Shakespeare, we trained an LSTM comprising one embedding layer, one LSTM layer, and one fully connected layer. This model structure originates from the Leaf database\cite{caldas2018leaf}.

\noindent \textbf{Baselines and Implementation Details.} We compared the test accuracies of FedCAda and the following federated learning algorithms: Fedavg, Fedadam, Fedams, FedSps, and Fafed, and reported the optimal performances of these algorithms' global models. For Fedadam and Fedams, we set $\beta_1$ to $0.9$, $\beta_2$ to $0.99$, and meticulously tuned the global learning rate $\eta_g$. For FedSps, we tuned parameters $c$ and $\gamma$ to achieve the best performance. As for our FedCAda and Fafed, the client-side $\beta_1$ was selected from the range ${0.1, 0.9}$, $\beta_2$ was set to $0.99$, and we carefully tuned the local learning rate $\eta_l$. We concurrently conducted experiments with FedCAda and all baselines under both cross-silo and cross-device settings. For the cross-silo setting, we had $20$ clients and set the select ratio per round $\rho$ to $1$. As for the cross-device setting, we had $100$ clients and set the select ratio per round $\rho$ to $0.2$. On each client, we allocated $75\%$ of the data for training and used the remaining $25\%$ for evaluation. By default, we ran $200$ rounds with the number of local epochs set to $3$.In the Shakespeare dataset, we set the number of local epochs to $1$. The global test set at the server side is a collection of data from 20 clients selected after removing the selected clients from the original dataset.

\begin{table*}[htbp]
\caption{Top-1 test accuracy (\%) achieved by comparing FL baseline methods and FedCAda on CIFAR-10, FMNIST, and Shakespeare three datasets under Cross-Silo and Cross-Device settings with $E = 3$ and $T = 200$. The bold fonts denote the best approach and the underlined ones are the second-best performance.}
\begin{tabular*}{\linewidth}{@{\extracolsep{\fill}} p{2.3 cm}|ccccc|ccccc}
\toprule
\rule{0pt}{2.5ex} Setting & \multicolumn{5}{c|}{Cross-Silo (20 clients, select ratio $\rho = 1$)}                            & \multicolumn{5}{c}{Cross-Device (100 clients, select ratio $\rho = 0.2$)}                       \\ \midrule
\multirow{2}{*}{Dataset} & \multicolumn{2}{c}{CIFAR-10} & \multicolumn{2}{c}{FMNIST} & Shakespeare & \multicolumn{2}{c}{CIFAR-10} & \multicolumn{2}{c}{FMNIST} & Shakespeare\\
\cmidrule(r){2-3} \cmidrule(r){4-5}  \cmidrule(r){6-6} \cmidrule(r){7-8} \cmidrule(r){9-10} \cmidrule(r){11-11} & IID & \makecell{NonIID \\ ($\alpha=0.1$)} & IID & \makecell{NonIID \\ ($\alpha=0.1$)} & NonIID & IID & \makecell{NonIID \\ ($\alpha=0.1$)} & IID & \makecell{NonIID \\ ($\alpha=0.1$)} & NonIID\\ \midrule
FedAvg \cite{mcmahan2017communication} & 65.79 & 59.58 & 89.16 & 85.38 & \underline{45.21} &  56.71   & 53.34 & \underline{88.18} & 81.53 & 48.47\\
FedAdam \cite{reddi2020adaptive} & 64.32 & 59.40 & \textbf{89.20} & 85.29 &  43.18  &  57.29   & \textbf{59.06}  & 87.40 & 82.60 & 49.60 \\
FedAMS \cite{wang2022communication} & 62.51 & \underline{61.19} & 89.14 & 85.72 &  43.01  &  56.00   & 56.06 & 87.80 & 82.50 &49.56\\
FedSps \cite{mukherjee2023locally} & 64.43 & 60.39 & 89.09 & 82.58 &  43.29 & \underline{63.04} & 53.79 & 86.71 & 76.78 & 55.50\\
FAFed \cite{wu2023faster} & \underline{66.07} & 61.15 & \underline{89.18} & \underline{86.15} & 43.03 &  62.70   & 57.97 & \textbf{88.29} & \underline{82.91} & \underline{55.74} \\
FedCAda (Ours) & \textbf{67.24} & \textbf{62.52} & 89.11 & \textbf{86.35} & \textbf{45.84}  &  \textbf{66.508}   & \underline{58.26} & 88.17 & \textbf{84.82} & \textbf{55.79}\\ \bottomrule
\end{tabular*}
\label{tableacc}
\end{table*}

\subsection{Performance Evaluation}
\textbf{METRIC: ACC(Accuracy).}
The variable acc represents the accuracy of the model on a given dataset. Accuracy is a crucial performance metric for evaluating classification models, indicating the proportion of correctly classified samples by the model. By calculating accuracy, we can assess the classification performance of the model on the given dataset. Higher accuracy implies a stronger classification ability of the model. In CIFAR-10 and FashionMNIST databases, we employed acc as the evaluation metric. 

\subsubsection{Main Results}

In this section, we evaluate the efficacy of FedCAda in two distinct settings: cross-silo and cross-device, across three datasets. This evaluation considers both IID and Non-IID data configurations.

Figure \ref{fmnistacc} illustrates the comparison between our algorithm and other methods on train loss and global test accuracy in two scenarios: cross-silo and cross-device, on the CIFAR-10 dataset. In the cross-silo scenario, it can be observed that both our method and FaFed outperform other methods significantly in terms of convergence speed and accuracy. However, the differences in adjustment of $m$ and $v$ between our method and FaFed led to our method outperforming FaFed in the final results as well. In the cross-device scenario, although the performance of FedCAda and FaFed does not significantly outpace other methods, it can still be observed that FedCAda and FaFed consistently remain at the top of the Global Model ACC curve, with FedCAda yielding better results than FaFed.

In Table \ref{tableacc}, in addition to the final accuracy on CIFAR-10, we also present the results on the FMNIST dataset and the Shakespeare dataset for both cross-silo and cross-device scenarios, as well as the methods for cross-silo and cross-device when the data on both datasets are divided IID. In the non-IID scenario, for FMNIST, both FaFed and FedCAda perform well on CIFAR-10, with FedCAda outperforming FaFed. Under IID data distribution, in the cross-silo scenario, FaFed and FedCAda achieve the best performance, whereas, in the cross-device scenario of CIFAR-10, FedCAda remains the best-performing method, but FaFed's results are surpassed by FedSps. However, under IID distribution on the FMNIST dataset, the global test accuracy among different methods becomes very close, and FedCAda and FaFed no longer maintain the best performance. In the Shakespeare dataset, when we select fewer than $20$ clients, the performance of the methods chosen for comparison is inferior to FedAvg. However, when the number of clients increases to 100, the accuracy of all methods significantly rises, surpassing FedAvg.

\begin{figure}
  \centering
  \begin{subfigure}[b]{\linewidth} 
    \includegraphics[width=\textwidth]{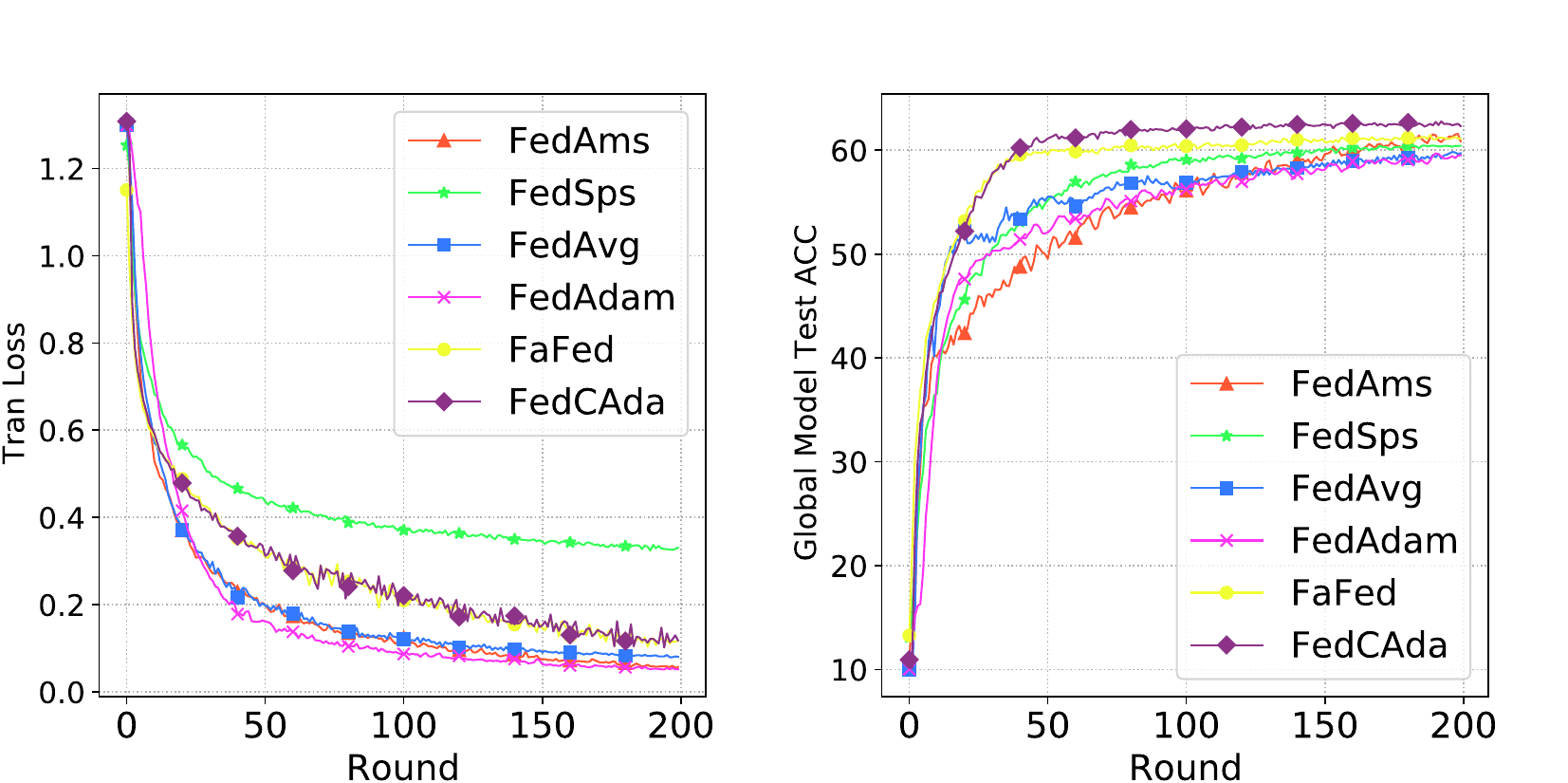}
    \caption{Cross-Silo (20 clients, select ratio $\rho = 1$)}
    \label{fig:sub1}
  \end{subfigure}
  \vspace{0.3cm}
  \begin{subfigure}[b]{\linewidth} 
    \includegraphics[width=\textwidth]{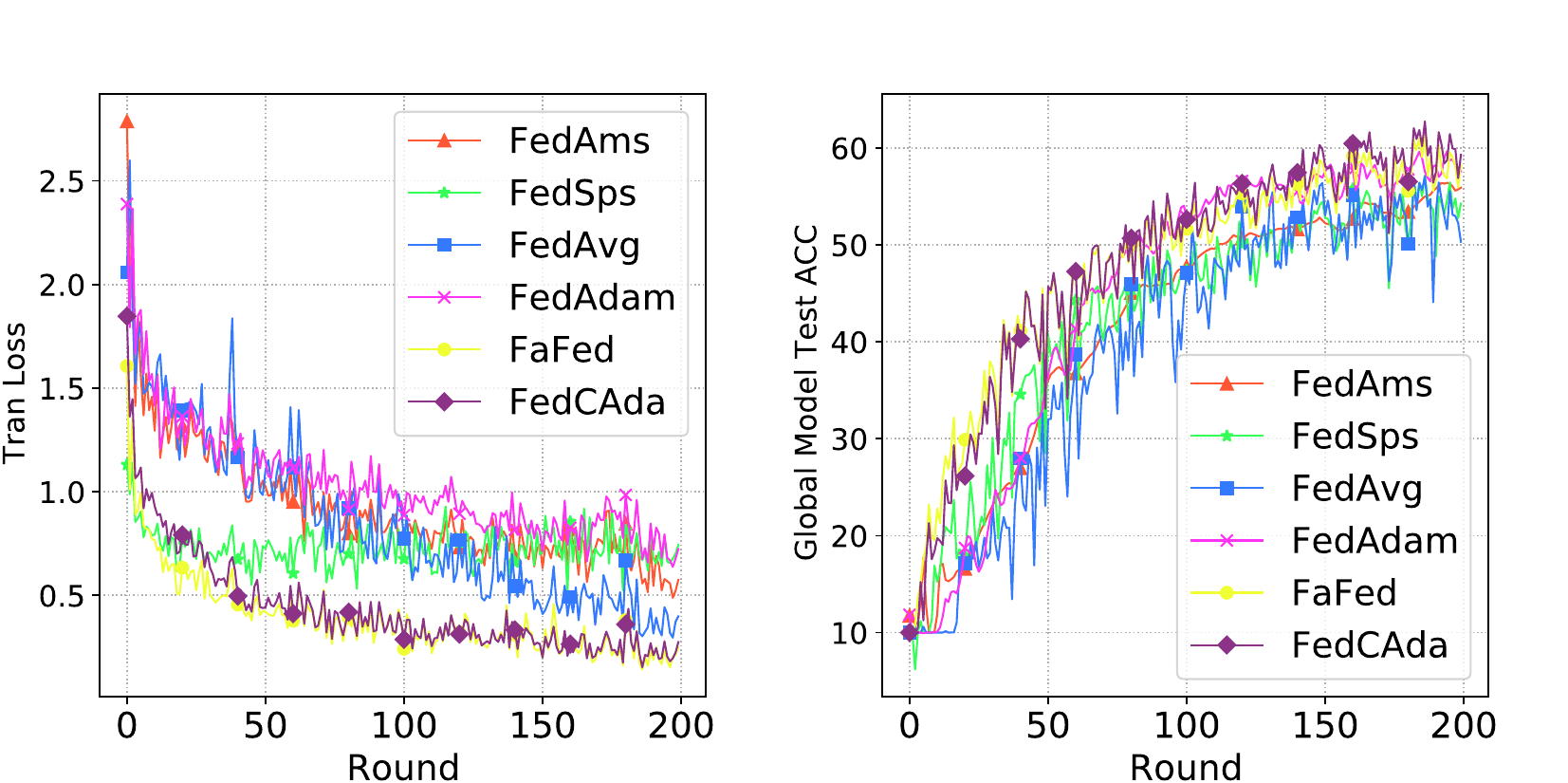}
    \caption{Cross-Device (100 clients, select ratio $\rho = 0.2$)}
    \label{fig:sub2}
  \end{subfigure}
  \caption{The training loss (left) and global model test accuracy (right) curves for FedCAda and other federated learning baselines on training CIFAR-10 with $E=3$ and $T=200$.}
  \label{fig:sub2}
\label{cifaracc}
\end{figure}

\begin{figure}
  \centering
  \begin{subfigure}[b]{\linewidth} 
    \includegraphics[width=\textwidth]{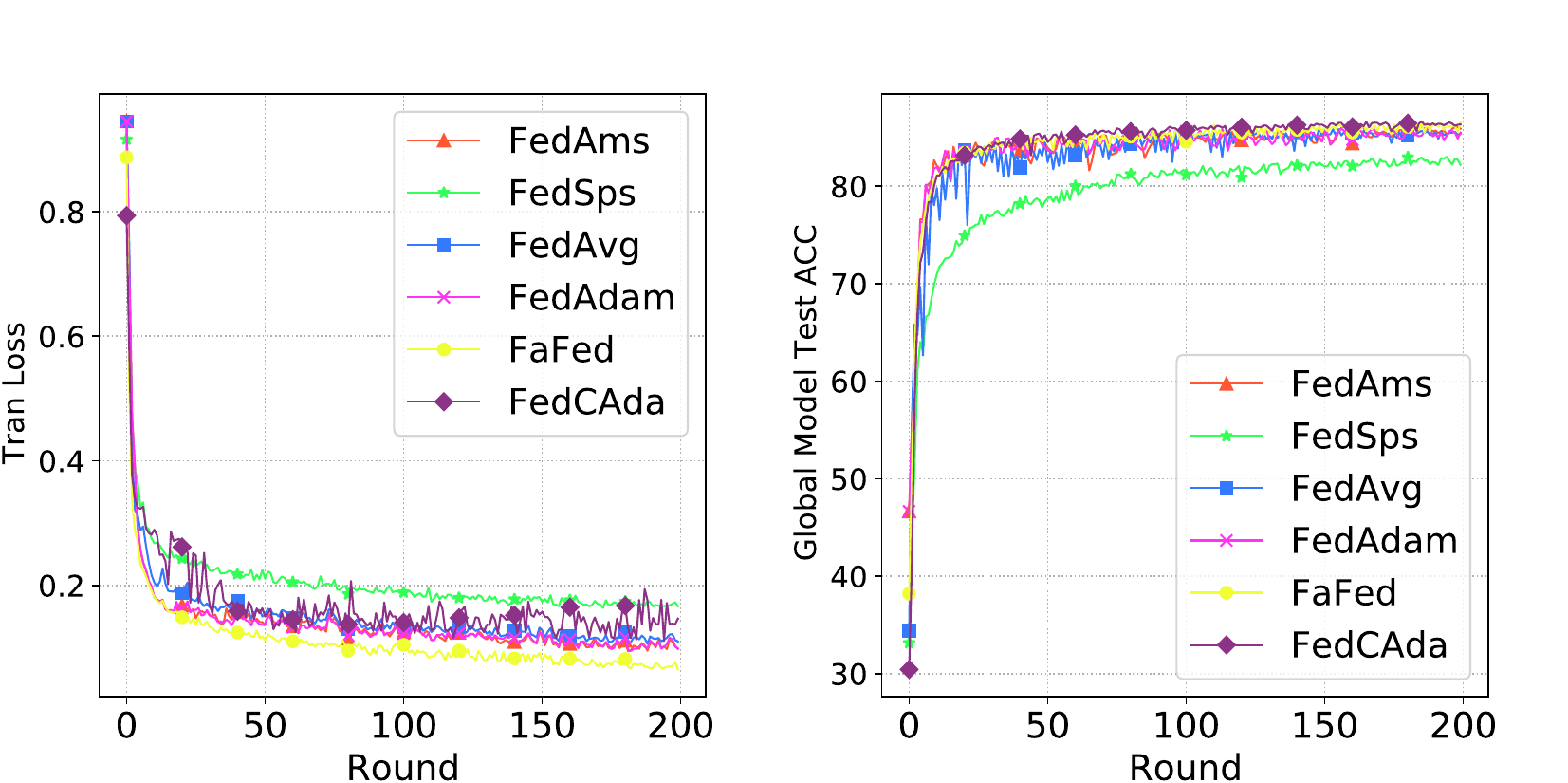}
    \caption{Cross-Silo (20 clients, select ratio $\rho = 1$)}
    \label{fig:sub1}
  \end{subfigure}

  \vspace{0.3cm}

  \begin{subfigure}[b]{\linewidth} 
    \includegraphics[width=\textwidth]{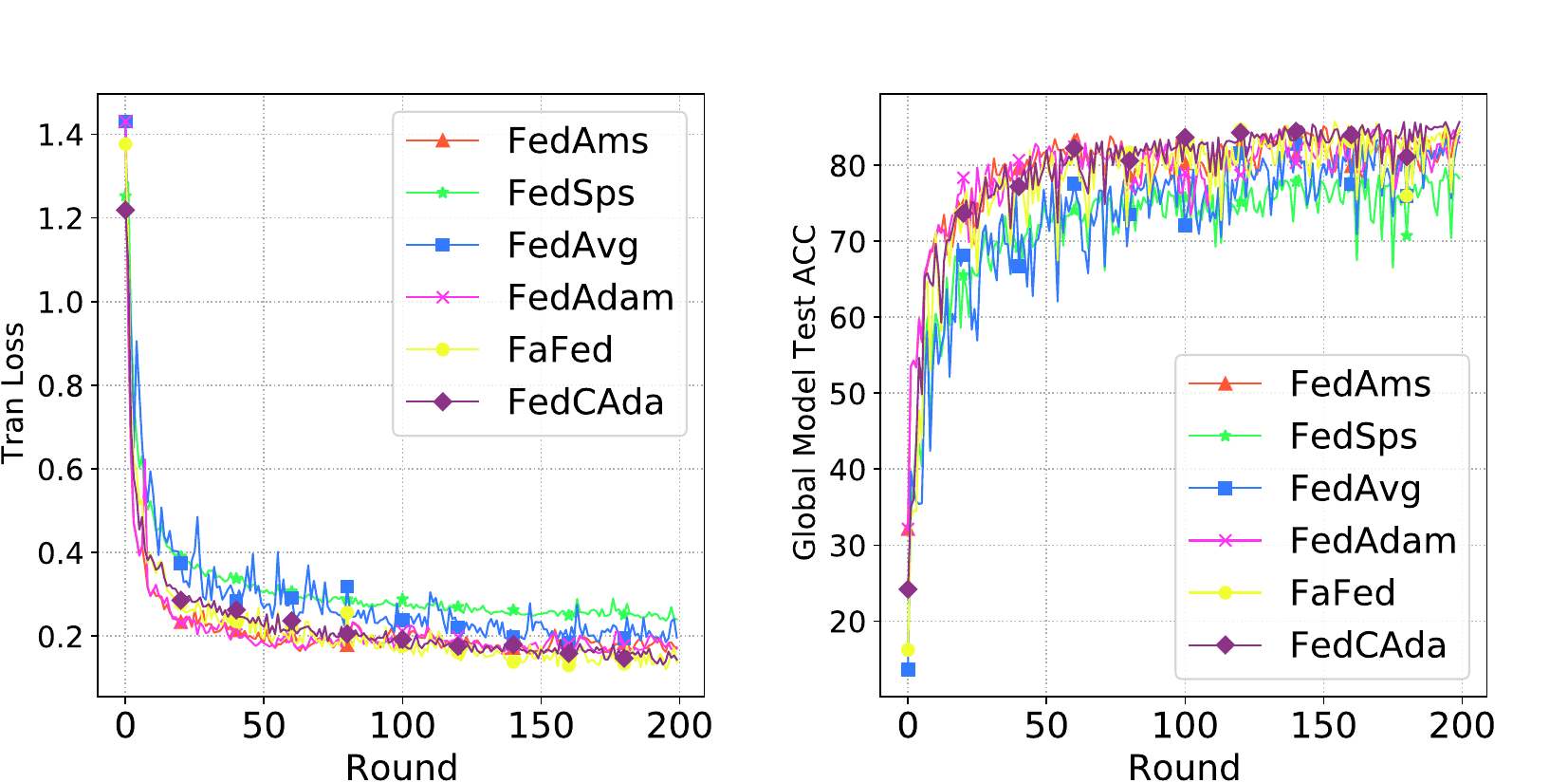}
    \caption{Cross-Device (100 clients, select ratio $\rho = 0.2$)}
    \label{fig:sub2}
  \end{subfigure}
  \caption{The training loss (left) and global model test accuracy (right) curves for FedCAda and other federated learning baselines on training FashionMNIST with $E=3$ and $T=200$.}
  \label{fig:sub2}
\label{fmnistacc}
\end{figure}

\subsubsection{Results for Different Model Architectures}
In this section, we compare the performance of FedCAda across different model architectures. Specifically, we compare the performance of using SimpleCNN and ResNet-18 \cite{he2016deep} models on CIFAR-10, under a cross-device setting and Non-IID data distribution. 

The experimental results presented in Table \ref{resnet} demonstrate the performance of various federated learning models across two neural network architectures. Based on the experimental findings, the transition from the SimlpeCNN to the ResNet-18 model resulted in a general reduction in accuracy (acc). Despite this decline, it is notable that the majority of the methods exhibited marked improvements in comparison to FedAvg. Specifically, following the shift to the ResNet-18 model, FedCAda showcased the highest accuracy among all methods, thereby sustaining superior performance. This performance superiority was maintained despite the decrease in accuracy in comparison to the SimpleCNN model. Conversely, it is important to highlight that FedSps demonstrated a lower accuracy when compared to FedAvg. These observations underscore the nuanced impact of model transitions on the performance of different methods, thus calling for further investigation and analysis.

\begin{table*}[htbp]
\caption{The performance of compared methods with different model architectures on training Non-IID CIFAR-10 dataset in cross-device setting.}
\label{resnet}
\begin{tabular}{l|cccccc}
\toprule
Model     & FedAvg & FedAdam & FedAMS & FedSps & FAFed & FedCAda \\ \midrule
SimpleCNN & 53.34  & 59.06   & 56.06  & 53.79  & 57.97 & 58.26   \\
ResNet-18 & 49.86  & 54.39   & 53.81  & 44.44  & 56.75 & 61.32   \\
\bottomrule
\end{tabular}
\end{table*}

\begin{figure}[htbp]
\centering
\includegraphics[width=\linewidth]{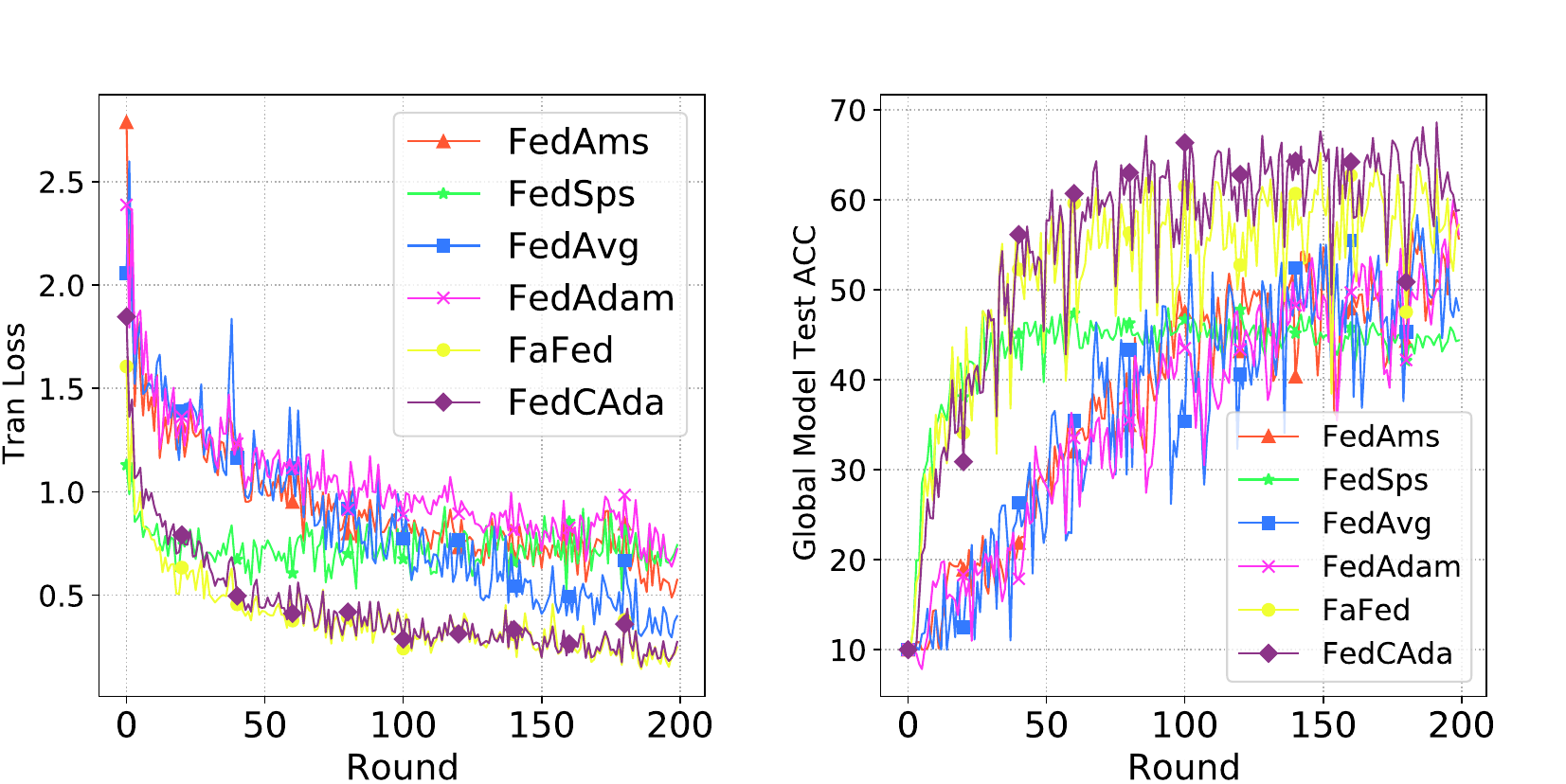}
\caption{The training loss (left) and global model test accuracy (right) curves for FedCAda and other federated learning baselines on training CIFAR-10 with ResNet-18 model in cross-silo setting.}
\label{figresnet}
\end{figure}

\subsection{Ablation Study}

In this section, we aim to address two aspects of our proposed idea through ablation experiments:

\begin{enumerate}
    \item In this study, we present a novel adaptation to the method of adjusting the variables $m$ and $v$ within the adaptive learning rate framework. Specifically, we propose the modification of the original subtraction operation in the denominator to an addition operation. This alteration leads to a decrease in the magnitude of the values of $m$ and $v$. By implementing this change, the square root operation inherent in the calculation of the adaptive learning rate effectively mitigates any increase in the learning rate of the model. As a result, the magnitude of the model's update step on the client side is reduced, facilitating more conservative updates. This modification addresses and improves the convergence issues attributed to client drift, while concurrently enhancing the model's convergence rate. Further investigation will be conducted to examine the implications of not employing the $m$ and $v$ adjustment mechanism.
    \item Additionally, our algorithm introduces a novel approach by modifying the subtraction operation in the denominator of the $m$ and $v$ adjustment process to an addition, transforming the denominator into a value of $1$ plus a number that incrementally approaches $0$ as $T$ advances. This study aimed to determine whether approaching zero at different rates would have different effects. For this comparison, we have chosen three alternative functions—exponential, power, and sine functions as counterparts to the original formulation. 
\end{enumerate}

Consequently, this segment of the ablation experiments is dedicated to evaluating two critical aspects:

\begin{enumerate}
\item A comparison between our newly designed algorithm and scenarios in which the client side does not implement $m$ and $v$ adjustment.
\item An analysis of different forms of denominators during the adjustment process, with a focus on three alternative functions—exponential, power, and trigonometric functions—as mentioned in the methodology section, contrasted against the original formulation.
\end{enumerate}

\begin{table}[ht]
\centering
\caption{Performance comparison under FedCAda with different adjustment functions on training Non-IID CIFAR-10 dataset in cross-silo setting. FedCAda w/o functions means without $m$ and $v$ adjustment process}
\begin{tabularx}{0.9\linewidth}{|>{\arraybackslash}X|>{\centering\arraybackslash}X|}
\hline
Adjustment Function & Global Model Test ACC \\ \hline
FedCAda w/o functions  & 57.29 \\ \hline
FedCAda w/ $(1+\beta^t)$ & 62.52 \\ \hline
FedCAda w/ $(1+(\beta^t)^2)$ & 63.40 \\ \hline
FedCAda w/ $(1+\sin(\beta^t))$ & 62.77 \\ \hline
FedCAda w/ $(1+\sqrt{\beta^t})$ & 62.48 \\ \hline
\end{tabularx}
\label{difffunc}
\end{table}

Through an analysis of the initial two rows in Table \ref{difffunc}, it is apparent that the utilization of adjustment functions greatly improves FedCAda's performance. Upon reviewing the remaining four rows, we can confirm that the adjustment functions currently implemented by FedCAda are already quite efficient. Of the four functions offered, the exponential function function yields marginally superior results to FedCAda.

\section{Conclusion}

In conclusion, this paper introduces a new federated client adaptive algorithm called FedCAda, which aims to address the challenge of balancing acceleration and stability in federated learning. The proposed approach prioritizes speeding up the convergence on the client-side while ensuring the overall stability and performance of FL algorithms. To achieve this, FedCAda uses the Adam algorithm within the clients' framework in FL, allowing it to dynamically adjust the correction process of $m$ and $v$ on the client-side and aggregate adaptive algorithm parameters on the server-side.

We conducted extensive experiments on CV and NLP datasets and demonstrated that FedCAda outperforms or closely matches state-of-the-art adaptive methods in terms of adaptability, convergence, stability, and overall performance on the CIFAR-10, FMNIST, and Shakespeare datasets. Furthermore, our investigation into modifying the denominator during the adjustment process sheds light on the effectiveness of different adjustment approaches, indicating the efficacy of our proposed method.

\begin{acks}
To Robert, for the bagels and explaining CMYK and color spaces.
\end{acks}


\bibliographystyle{ACM-Reference-Format}
\bibliography{sample-base}


\begin{thebibliography}{35}


\ifx \showCODEN    \undefined \def \showCODEN     #1{\unskip}     \fi
\ifx \showDOI      \undefined \def \showDOI       #1{#1}\fi
\ifx \showISBNx    \undefined \def \showISBNx     #1{\unskip}     \fi
\ifx \showISBNxiii \undefined \def \showISBNxiii  #1{\unskip}     \fi
\ifx \showISSN     \undefined \def \showISSN      #1{\unskip}     \fi
\ifx \showLCCN     \undefined \def \showLCCN      #1{\unskip}     \fi
\ifx \shownote     \undefined \def \shownote      #1{#1}          \fi
\ifx \showarticletitle \undefined \def \showarticletitle #1{#1}   \fi
\ifx \showURL      \undefined \def \showURL       {\relax}        \fi
\providecommand\bibfield[2]{#2}
\providecommand\bibinfo[2]{#2}
\providecommand\natexlab[1]{#1}
\providecommand\showeprint[2][]{arXiv:#2}

\bibitem[Brown et~al\mbox{.}(2020)]%
        {brown2020language}
\bibfield{author}{\bibinfo{person}{Tom Brown}, \bibinfo{person}{Benjamin Mann}, \bibinfo{person}{Nick Ryder}, \bibinfo{person}{Melanie Subbiah}, \bibinfo{person}{Jared~D Kaplan}, \bibinfo{person}{Prafulla Dhariwal}, \bibinfo{person}{Arvind Neelakantan}, \bibinfo{person}{Pranav Shyam}, \bibinfo{person}{Girish Sastry}, \bibinfo{person}{Amanda Askell}, {et~al\mbox{.}}} \bibinfo{year}{2020}\natexlab{}.
\newblock \showarticletitle{Language models are few-shot learners}.
\newblock \bibinfo{journal}{\emph{Advances in neural information processing systems}}  \bibinfo{volume}{33} (\bibinfo{year}{2020}), \bibinfo{pages}{1877--1901}.
\newblock


\bibitem[Caldarola et~al\mbox{.}(2022)]%
        {caldarola2022improving}
\bibfield{author}{\bibinfo{person}{Debora Caldarola}, \bibinfo{person}{Barbara Caputo}, {and} \bibinfo{person}{Marco Ciccone}.} \bibinfo{year}{2022}\natexlab{}.
\newblock \showarticletitle{Improving generalization in federated learning by seeking flat minima}. In \bibinfo{booktitle}{\emph{European Conference on Computer Vision}}. Springer, \bibinfo{pages}{654--672}.
\newblock


\bibitem[Caldas et~al\mbox{.}(2018)]%
        {caldas2018leaf}
\bibfield{author}{\bibinfo{person}{Sebastian Caldas}, \bibinfo{person}{Sai Meher~Karthik Duddu}, \bibinfo{person}{Peter Wu}, \bibinfo{person}{Tian Li}, \bibinfo{person}{Jakub Kone{\v{c}}n{\`y}}, \bibinfo{person}{H~Brendan McMahan}, \bibinfo{person}{Virginia Smith}, {and} \bibinfo{person}{Ameet Talwalkar}.} \bibinfo{year}{2018}\natexlab{}.
\newblock \showarticletitle{Leaf: A benchmark for federated settings}.
\newblock \bibinfo{journal}{\emph{arXiv preprint arXiv:1812.01097}} (\bibinfo{year}{2018}).
\newblock


\bibitem[Chen et~al\mbox{.}(2023)]%
        {chen2023convergence}
\bibfield{author}{\bibinfo{person}{Xiangyi Chen}, \bibinfo{person}{Belhal Karimi}, \bibinfo{person}{Weijie Zhao}, {and} \bibinfo{person}{Ping Li}.} \bibinfo{year}{2023}\natexlab{}.
\newblock \showarticletitle{On the convergence of decentralized adaptive gradient methods}. In \bibinfo{booktitle}{\emph{Asian Conference on Machine Learning}}. PMLR, \bibinfo{pages}{217--232}.
\newblock


\bibitem[Dosovitskiy et~al\mbox{.}(2020)]%
        {dosovitskiy2020image}
\bibfield{author}{\bibinfo{person}{Alexey Dosovitskiy}, \bibinfo{person}{Lucas Beyer}, \bibinfo{person}{Alexander Kolesnikov}, \bibinfo{person}{Dirk Weissenborn}, \bibinfo{person}{Xiaohua Zhai}, \bibinfo{person}{Thomas Unterthiner}, \bibinfo{person}{Mostafa Dehghani}, \bibinfo{person}{Matthias Minderer}, \bibinfo{person}{Georg Heigold}, \bibinfo{person}{Sylvain Gelly}, {et~al\mbox{.}}} \bibinfo{year}{2020}\natexlab{}.
\newblock \showarticletitle{An image is worth 16x16 words: Transformers for image recognition at scale}.
\newblock \bibinfo{journal}{\emph{arXiv preprint arXiv:2010.11929}} (\bibinfo{year}{2020}).
\newblock


\bibitem[Duchi et~al\mbox{.}(2011)]%
        {duchi2011adaptive}
\bibfield{author}{\bibinfo{person}{John Duchi}, \bibinfo{person}{Elad Hazan}, {and} \bibinfo{person}{Yoram Singer}.} \bibinfo{year}{2011}\natexlab{}.
\newblock \showarticletitle{Adaptive subgradient methods for online learning and stochastic optimization.}
\newblock \bibinfo{journal}{\emph{Journal of machine learning research}} \bibinfo{volume}{12}, \bibinfo{number}{7} (\bibinfo{year}{2011}).
\newblock


\bibitem[Goodfellow et~al\mbox{.}(2020)]%
        {goodfellow2020generative}
\bibfield{author}{\bibinfo{person}{Ian Goodfellow}, \bibinfo{person}{Jean Pouget-Abadie}, \bibinfo{person}{Mehdi Mirza}, \bibinfo{person}{Bing Xu}, \bibinfo{person}{David Warde-Farley}, \bibinfo{person}{Sherjil Ozair}, \bibinfo{person}{Aaron Courville}, {and} \bibinfo{person}{Yoshua Bengio}.} \bibinfo{year}{2020}\natexlab{}.
\newblock \showarticletitle{Generative adversarial networks}.
\newblock \bibinfo{journal}{\emph{Commun. ACM}} \bibinfo{volume}{63}, \bibinfo{number}{11} (\bibinfo{year}{2020}), \bibinfo{pages}{139--144}.
\newblock


\bibitem[He et~al\mbox{.}(2016)]%
        {he2016deep}
\bibfield{author}{\bibinfo{person}{Kaiming He}, \bibinfo{person}{Xiangyu Zhang}, \bibinfo{person}{Shaoqing Ren}, {and} \bibinfo{person}{Jian Sun}.} \bibinfo{year}{2016}\natexlab{}.
\newblock \showarticletitle{Deep residual learning for image recognition}. In \bibinfo{booktitle}{\emph{Proceedings of the IEEE conference on computer vision and pattern recognition}}. \bibinfo{pages}{770--778}.
\newblock


\bibitem[Jin et~al\mbox{.}(2023)]%
        {jin2023feddyn}
\bibfield{author}{\bibinfo{person}{Cheng Jin}, \bibinfo{person}{Xuandong Chen}, \bibinfo{person}{Yi Gu}, {and} \bibinfo{person}{Qun Li}.} \bibinfo{year}{2023}\natexlab{}.
\newblock \showarticletitle{FedDyn: A dynamic and efficient federated distillation approach on Recommender System}. In \bibinfo{booktitle}{\emph{2022 IEEE 28th International Conference on Parallel and Distributed Systems (ICPADS)}}. IEEE, \bibinfo{pages}{786--793}.
\newblock


\bibitem[Kairouz et~al\mbox{.}(2021)]%
        {kairouz2021advances}
\bibfield{author}{\bibinfo{person}{Peter Kairouz}, \bibinfo{person}{H~Brendan McMahan}, \bibinfo{person}{Brendan Avent}, \bibinfo{person}{Aur{\'e}lien Bellet}, \bibinfo{person}{Mehdi Bennis}, \bibinfo{person}{Arjun~Nitin Bhagoji}, \bibinfo{person}{Kallista Bonawitz}, \bibinfo{person}{Zachary Charles}, \bibinfo{person}{Graham Cormode}, \bibinfo{person}{Rachel Cummings}, {et~al\mbox{.}}} \bibinfo{year}{2021}\natexlab{}.
\newblock \showarticletitle{Advances and open problems in federated learning}.
\newblock \bibinfo{journal}{\emph{Foundations and trends{\textregistered} in machine learning}} \bibinfo{volume}{14}, \bibinfo{number}{1--2} (\bibinfo{year}{2021}), \bibinfo{pages}{1--210}.
\newblock


\bibitem[Karimireddy et~al\mbox{.}(2020)]%
        {karimireddy2020scaffold}
\bibfield{author}{\bibinfo{person}{Sai~Praneeth Karimireddy}, \bibinfo{person}{Satyen Kale}, \bibinfo{person}{Mehryar Mohri}, \bibinfo{person}{Sashank Reddi}, \bibinfo{person}{Sebastian Stich}, {and} \bibinfo{person}{Ananda~Theertha Suresh}.} \bibinfo{year}{2020}\natexlab{}.
\newblock \showarticletitle{Scaffold: Stochastic controlled averaging for federated learning}. In \bibinfo{booktitle}{\emph{International conference on machine learning}}. PMLR, \bibinfo{pages}{5132--5143}.
\newblock


\bibitem[Kingma and Ba(2014)]%
        {kingma2014adam}
\bibfield{author}{\bibinfo{person}{Diederik~P Kingma} {and} \bibinfo{person}{Jimmy Ba}.} \bibinfo{year}{2014}\natexlab{}.
\newblock \showarticletitle{Adam: A method for stochastic optimization}.
\newblock \bibinfo{journal}{\emph{arXiv preprint arXiv:1412.6980}} (\bibinfo{year}{2014}).
\newblock


\bibitem[Kornblith et~al\mbox{.}(2019)]%
        {kornblith2019similarity}
\bibfield{author}{\bibinfo{person}{Simon Kornblith}, \bibinfo{person}{Mohammad Norouzi}, \bibinfo{person}{Honglak Lee}, {and} \bibinfo{person}{Geoffrey Hinton}.} \bibinfo{year}{2019}\natexlab{}.
\newblock \showarticletitle{Similarity of neural network representations revisited}. In \bibinfo{booktitle}{\emph{International conference on machine learning}}. PMLR, \bibinfo{pages}{3519--3529}.
\newblock


\bibitem[Krizhevsky et~al\mbox{.}(2009)]%
        {krizhevsky2009learning}
\bibfield{author}{\bibinfo{person}{Alex Krizhevsky}, \bibinfo{person}{Geoffrey Hinton}, {et~al\mbox{.}}} \bibinfo{year}{2009}\natexlab{}.
\newblock \showarticletitle{Learning multiple layers of features from tiny images}.
\newblock  (\bibinfo{year}{2009}).
\newblock


\bibitem[Li et~al\mbox{.}(2021)]%
        {li2021model}
\bibfield{author}{\bibinfo{person}{Qinbin Li}, \bibinfo{person}{Bingsheng He}, {and} \bibinfo{person}{Dawn Song}.} \bibinfo{year}{2021}\natexlab{}.
\newblock \showarticletitle{Model-contrastive federated learning}. In \bibinfo{booktitle}{\emph{Proceedings of the IEEE/CVF conference on computer vision and pattern recognition}}. \bibinfo{pages}{10713--10722}.
\newblock


\bibitem[Li et~al\mbox{.}(2020)]%
        {li2020federated}
\bibfield{author}{\bibinfo{person}{Tian Li}, \bibinfo{person}{Anit~Kumar Sahu}, \bibinfo{person}{Ameet Talwalkar}, {and} \bibinfo{person}{Virginia Smith}.} \bibinfo{year}{2020}\natexlab{}.
\newblock \showarticletitle{Federated learning: Challenges, methods, and future directions}.
\newblock \bibinfo{journal}{\emph{IEEE signal processing magazine}} \bibinfo{volume}{37}, \bibinfo{number}{3} (\bibinfo{year}{2020}), \bibinfo{pages}{50--60}.
\newblock


\bibitem[Li et~al\mbox{.}(2023)]%
        {li2023revisiting}
\bibfield{author}{\bibinfo{person}{Zexi Li}, \bibinfo{person}{Tao Lin}, \bibinfo{person}{Xinyi Shang}, {and} \bibinfo{person}{Chao Wu}.} \bibinfo{year}{2023}\natexlab{}.
\newblock \showarticletitle{Revisiting weighted aggregation in federated learning with neural networks}. In \bibinfo{booktitle}{\emph{International Conference on Machine Learning}}. PMLR, \bibinfo{pages}{19767--19788}.
\newblock


\bibitem[Lin et~al\mbox{.}(2020)]%
        {lin2020ensemble}
\bibfield{author}{\bibinfo{person}{Tao Lin}, \bibinfo{person}{Lingjing Kong}, \bibinfo{person}{Sebastian~U Stich}, {and} \bibinfo{person}{Martin Jaggi}.} \bibinfo{year}{2020}\natexlab{}.
\newblock \showarticletitle{Ensemble distillation for robust model fusion in federated learning}.
\newblock \bibinfo{journal}{\emph{Advances in Neural Information Processing Systems}}  \bibinfo{volume}{33} (\bibinfo{year}{2020}), \bibinfo{pages}{2351--2363}.
\newblock


\bibitem[Long et~al\mbox{.}(2023)]%
        {long2023fedcd}
\bibfield{author}{\bibinfo{person}{Yunfei Long}, \bibinfo{person}{Zhe Xue}, \bibinfo{person}{Lingyang Chu}, \bibinfo{person}{Tianlong Zhang}, \bibinfo{person}{Junjiang Wu}, \bibinfo{person}{Yu Zang}, {and} \bibinfo{person}{Junping Du}.} \bibinfo{year}{2023}\natexlab{}.
\newblock \showarticletitle{Fedcd: A classifier debiased federated learning framework for non-iid data}. In \bibinfo{booktitle}{\emph{Proceedings of the 31st ACM International Conference on Multimedia}}. \bibinfo{pages}{8994--9002}.
\newblock


\bibitem[Lu et~al\mbox{.}(2023)]%
        {lu2023federated}
\bibfield{author}{\bibinfo{person}{Jianghu Lu}, \bibinfo{person}{Shikun Li}, \bibinfo{person}{Kexin Bao}, \bibinfo{person}{Pengju Wang}, \bibinfo{person}{Zhenxing Qian}, {and} \bibinfo{person}{Shiming Ge}.} \bibinfo{year}{2023}\natexlab{}.
\newblock \showarticletitle{Federated Learning with Label-Masking Distillation}. In \bibinfo{booktitle}{\emph{Proceedings of the 31st ACM International Conference on Multimedia}}. \bibinfo{pages}{222--232}.
\newblock


\bibitem[McMahan et~al\mbox{.}(2017)]%
        {mcmahan2017communication}
\bibfield{author}{\bibinfo{person}{Brendan McMahan}, \bibinfo{person}{Eider Moore}, \bibinfo{person}{Daniel Ramage}, \bibinfo{person}{Seth Hampson}, {and} \bibinfo{person}{Blaise~Aguera y Arcas}.} \bibinfo{year}{2017}\natexlab{}.
\newblock \showarticletitle{Communication-efficient learning of deep networks from decentralized data}. In \bibinfo{booktitle}{\emph{Artificial intelligence and statistics}}. PMLR, \bibinfo{pages}{1273--1282}.
\newblock


\bibitem[Mukherjee et~al\mbox{.}(2023)]%
        {mukherjee2023locally}
\bibfield{author}{\bibinfo{person}{Sohom Mukherjee}, \bibinfo{person}{Nicolas Loizou}, {and} \bibinfo{person}{Sebastian~U Stich}.} \bibinfo{year}{2023}\natexlab{}.
\newblock \showarticletitle{Locally Adaptive Federated Learning via Stochastic Polyak Stepsizes}.
\newblock \bibinfo{journal}{\emph{arXiv preprint arXiv:2307.06306}} (\bibinfo{year}{2023}).
\newblock


\bibitem[Nguyen et~al\mbox{.}(2022)]%
        {nguyen2022begin}
\bibfield{author}{\bibinfo{person}{John Nguyen}, \bibinfo{person}{Kshitiz Malik}, \bibinfo{person}{Maziar Sanjabi}, {and} \bibinfo{person}{Michael Rabbat}.} \bibinfo{year}{2022}\natexlab{}.
\newblock \showarticletitle{Where to begin? exploring the impact of pre-training and initialization in federated learning}.
\newblock \bibinfo{journal}{\emph{arXiv preprint arXiv:2206.15387}}  \bibinfo{volume}{4} (\bibinfo{year}{2022}).
\newblock


\bibitem[Qi et~al\mbox{.}(2023)]%
        {qi2023cross}
\bibfield{author}{\bibinfo{person}{Zhuang Qi}, \bibinfo{person}{Lei Meng}, \bibinfo{person}{Zitan Chen}, \bibinfo{person}{Han Hu}, \bibinfo{person}{Hui Lin}, {and} \bibinfo{person}{Xiangxu Meng}.} \bibinfo{year}{2023}\natexlab{}.
\newblock \showarticletitle{Cross-Silo Prototypical Calibration for Federated Learning with Non-IID Data}. In \bibinfo{booktitle}{\emph{Proceedings of the 31st ACM International Conference on Multimedia}}. \bibinfo{pages}{3099--3107}.
\newblock


\bibitem[Qu et~al\mbox{.}(2022)]%
        {qu2022generalized}
\bibfield{author}{\bibinfo{person}{Zhe Qu}, \bibinfo{person}{Xingyu Li}, \bibinfo{person}{Rui Duan}, \bibinfo{person}{Yao Liu}, \bibinfo{person}{Bo Tang}, {and} \bibinfo{person}{Zhuo Lu}.} \bibinfo{year}{2022}\natexlab{}.
\newblock \showarticletitle{Generalized federated learning via sharpness aware minimization}. In \bibinfo{booktitle}{\emph{International conference on machine learning}}. PMLR, \bibinfo{pages}{18250--18280}.
\newblock


\bibitem[Reddi et~al\mbox{.}(2020)]%
        {reddi2020adaptive}
\bibfield{author}{\bibinfo{person}{Sashank Reddi}, \bibinfo{person}{Zachary Charles}, \bibinfo{person}{Manzil Zaheer}, \bibinfo{person}{Zachary Garrett}, \bibinfo{person}{Keith Rush}, \bibinfo{person}{Jakub Kone{\v{c}}n{\`y}}, \bibinfo{person}{Sanjiv Kumar}, {and} \bibinfo{person}{H~Brendan McMahan}.} \bibinfo{year}{2020}\natexlab{}.
\newblock \showarticletitle{Adaptive federated optimization}.
\newblock \bibinfo{journal}{\emph{arXiv preprint arXiv:2003.00295}} (\bibinfo{year}{2020}).
\newblock


\bibitem[Reddi et~al\mbox{.}(2019)]%
        {reddi2019convergence}
\bibfield{author}{\bibinfo{person}{Sashank~J Reddi}, \bibinfo{person}{Satyen Kale}, {and} \bibinfo{person}{Sanjiv Kumar}.} \bibinfo{year}{2019}\natexlab{}.
\newblock \showarticletitle{On the convergence of adam and beyond}.
\newblock \bibinfo{journal}{\emph{arXiv preprint arXiv:1904.09237}} (\bibinfo{year}{2019}).
\newblock


\bibitem[Tieleman and Hinton(2017)]%
        {tieleman2017divide}
\bibfield{author}{\bibinfo{person}{Tijmen Tieleman} {and} \bibinfo{person}{G Hinton}.} \bibinfo{year}{2017}\natexlab{}.
\newblock \showarticletitle{Divide the gradient by a running average of its recent magnitude. coursera: Neural networks for machine learning}.
\newblock \bibinfo{journal}{\emph{Technical report}} (\bibinfo{year}{2017}).
\newblock


\bibitem[Wang et~al\mbox{.}(2022)]%
        {wang2022communication}
\bibfield{author}{\bibinfo{person}{Yujia Wang}, \bibinfo{person}{Lu Lin}, {and} \bibinfo{person}{Jinghui Chen}.} \bibinfo{year}{2022}\natexlab{}.
\newblock \showarticletitle{Communication-efficient adaptive federated learning}. In \bibinfo{booktitle}{\emph{International Conference on Machine Learning}}. PMLR, \bibinfo{pages}{22802--22838}.
\newblock


\bibitem[Wu et~al\mbox{.}(2023)]%
        {wu2023faster}
\bibfield{author}{\bibinfo{person}{Xidong Wu}, \bibinfo{person}{Feihu Huang}, \bibinfo{person}{Zhengmian Hu}, {and} \bibinfo{person}{Heng Huang}.} \bibinfo{year}{2023}\natexlab{}.
\newblock \showarticletitle{Faster adaptive federated learning}. In \bibinfo{booktitle}{\emph{Proceedings of the AAAI Conference on Artificial Intelligence}}, Vol.~\bibinfo{volume}{37}. \bibinfo{pages}{10379--10387}.
\newblock


\bibitem[Xiao et~al\mbox{.}(2017)]%
        {xiao2017fashion}
\bibfield{author}{\bibinfo{person}{Han Xiao}, \bibinfo{person}{Kashif Rasul}, {and} \bibinfo{person}{Roland Vollgraf}.} \bibinfo{year}{2017}\natexlab{}.
\newblock \showarticletitle{Fashion-mnist: a novel image dataset for benchmarking machine learning algorithms}.
\newblock \bibinfo{journal}{\emph{arXiv preprint arXiv:1708.07747}} (\bibinfo{year}{2017}).
\newblock


\bibitem[Yurochkin et~al\mbox{.}(2019)]%
        {yurochkin2019bayesian}
\bibfield{author}{\bibinfo{person}{Mikhail Yurochkin}, \bibinfo{person}{Mayank Agarwal}, \bibinfo{person}{Soumya Ghosh}, \bibinfo{person}{Kristjan Greenewald}, \bibinfo{person}{Nghia Hoang}, {and} \bibinfo{person}{Yasaman Khazaeni}.} \bibinfo{year}{2019}\natexlab{}.
\newblock \showarticletitle{Bayesian nonparametric federated learning of neural networks}. In \bibinfo{booktitle}{\emph{International conference on machine learning}}. PMLR, \bibinfo{pages}{7252--7261}.
\newblock


\bibitem[Zhang et~al\mbox{.}(2023b)]%
        {zhang2023improving}
\bibfield{author}{\bibinfo{person}{Pengling Zhang}, \bibinfo{person}{Huibin Yan}, \bibinfo{person}{Wenhui Wu}, {and} \bibinfo{person}{Shuoyao Wang}.} \bibinfo{year}{2023}\natexlab{b}.
\newblock \showarticletitle{Improving Federated Person Re-Identification through Feature-Aware Proximity and Aggregation}. In \bibinfo{booktitle}{\emph{Proceedings of the 31st ACM International Conference on Multimedia}}. \bibinfo{pages}{2498--2506}.
\newblock


\bibitem[Zhang et~al\mbox{.}(2023a)]%
        {zhang2023cuing}
\bibfield{author}{\bibinfo{person}{Yuxuan Zhang}, \bibinfo{person}{Lei Liu}, {and} \bibinfo{person}{Li Liu}.} \bibinfo{year}{2023}\natexlab{a}.
\newblock \showarticletitle{Cuing without sharing: A federated cued speech recognition framework via mutual knowledge distillation}. In \bibinfo{booktitle}{\emph{Proceedings of the 31st ACM International Conference on Multimedia}}. \bibinfo{pages}{8781--8789}.
\newblock


\bibitem[Zhao et~al\mbox{.}(2018)]%
        {zhao2018federated}
\bibfield{author}{\bibinfo{person}{Yue Zhao}, \bibinfo{person}{Meng Li}, \bibinfo{person}{Liangzhen Lai}, \bibinfo{person}{Naveen Suda}, \bibinfo{person}{Damon Civin}, {and} \bibinfo{person}{Vikas Chandra}.} \bibinfo{year}{2018}\natexlab{}.
\newblock \showarticletitle{Federated learning with non-iid data}.
\newblock \bibinfo{journal}{\emph{arXiv preprint arXiv:1806.00582}} (\bibinfo{year}{2018}).
\newblock


\end{thebibliography}

\end{document}